\pdfoutput=1
\documentclass{article}
\usepackage{ijcai22}

\usepackage{times}
\usepackage{soul}
\usepackage{url}
\usepackage[hidelinks]{hyperref}
\usepackage[utf8]{inputenc}
\usepackage[small]{caption}
\usepackage{graphicx}
\usepackage{amsmath}
\usepackage{amsthm}
\usepackage{booktabs}
\usepackage{algorithm}
\usepackage{algorithmic}
\title{Responsibility: An Example-based Explainable AI approach via Training Process Inspection}

\author{
Faraz Khadivpour$^1$
\and
Arghasree Banerjee$^1$\and
Matthew Guzdial $^{1}$
\affiliations
$^1$University of Alberta\\
\emails
\{khadivpo, arghasre, guzdial\}@ualberta.ca,
}

\begin{document}

\maketitle

\begin{abstract}
Explainable Artificial Intelligence (XAI) methods are intended to help human users better understand the decision making of an AI agent. However, many modern XAI approaches are unintuitive to end users, particularly those without prior AI or ML knowledge. In this paper we present a novel XAI approach we call Responsibility that identifies the most responsible training example for a particular decision. This example can then be shown as an explanation: ``this is what I (the AI) learned that led me to do that''. We present experimental results across a number of domains along with the results of an Amazon Mechanical Turk user study, comparing responsibility and existing XAI methods on an image classification task. Our results demonstrate that responsibility can help improve accuracy for both human end users and secondary ML models.
\end{abstract}

\section{Introduction}

Explainable artificial intelligence (XAI) is a research area that aims to develop approaches to make AI systems more understandable to human users \cite{biran2017explanation}.
Researchers have developed many XAI approaches to produce explanations, however the majority of these make use of post-hoc analysis, attempting to induce an explanation from observing the behavior of a final, trained model \cite{adadi2018peeking}. 
In addition, it is challenging to design XAI systems that are able to provide accurate explanations which are understandable to non-expert end users \cite{mohseni2021multidisciplinary}. 
The approaches that do meet the above two criteria are often limited to only a single application domain, such as image classification tasks \cite{vilone2020explainable}.
Thus, despite the large number of existing XAI approaches we still require novel, general methods that create human-understandable explanations to end-users. 

There are existing XAI approaches with greater generality, which can be applied to a variety of problems.
These include Local Interpretable model-agnostic explanations (LIME) \cite{ribeiro2016should}, Shapley additive explanations (SHAP)  \cite{lundberg2017unified}, Grad-CAM \cite{selvaraju2017grad}, and Influence Functions \cite{koh2017understanding} which altogether represent some of the most popular modern XAI approaches. 
While each of these methods have been evaluated under a number of tasks, they still largely rely on inductive processes and each was designed for some particular domain or type of input. 

In this paper, we present a novel XAI approach focused on a direct examination of the training process, rather than post-hoc induction.
The basic idea of the approach is to attempt to answer the question: ``What did the AI learn that led it to make that decision?''. \cite{cook2019framing}.
To accomplish this, we detect the training example that is the most \emph{responsible} for a machine learning (ML) model’s decision at inference time.
There have been a variety of prior approaches that attempt to identify the most influential training samples, including Influence functions \cite{koh2017understanding} and Shapely values \cite{ghorbani2019data}. These approaches have been referred to as both example-based and instance-based methods.
However, all of these attempt to identify something objective about the training data, regardless of what final knowledge a model derived during a particular training process. 
Given the stochastic nature of model training, it may not be that these most influential training samples were the most causal of a particular final model. 
In comparison, we observe each training process to determine what examples impacted the model, and what kind of impacts they had. 
Thus, we focus on the effects of a particular training process on a final model.
We argue that this allows our approach to more accurately answer this question of responsibility, in comparison to prior approaches.

We call our approach \emph{Responsibility}, as in the process by which we identify the most responsible training example(s) for a particular decision at inference time.
This differs from responsible AI, which is a methodology for the implementation of AI approaches in real organizations in regards to fairness, transparency, accountability, and safety \cite{arrieta2020explainable}.
It is designed to be general across problem domains, with a focus on non-expert end users.

Our contributions in this work are as follows. 
First, we present a definition of our novel XAI approach \emph{Responsibility}.
Second, we propose a quantitative evaluation for XAI approaches in terms of evaluator accuracy, as an approximation for human understanding.
Third, we present the results of an Amazon Mechanical Turk human subject study comparing four popular XAI approaches and Responsibility on an image classification task. 
In addition, we include a number of standard XAI experiments in the technical appendix to demonstrate how Responsibility can help us better understand particular training processes.
Taken together, these results indicate that Responsibility is a novel XAI approach that is beneficial to end users in terms of performance and understanding of the trained model.

\section{Related work}

The majority of XAI approaches have typically been limited to a particular type of problem and data.
Vilone and Longo conducted a systemic review of 196 XAI articles and found that 80\% of these only worked for classification problems \cite{vilone2020explainable}.
Further, they determined that nearly half of the approaches were limited to a single type of data.
Our approach differs from prior work in terms of allowing for any deep neural network training problem.
In return, developers must make choices in terms of the selection of what parameters of the network to focus on, which we discuss further below.

A large amount of prior XAI work focuses on image classification and relies on the concept of saliency in images, as initially presented in \cite{itti1998model}.
Saliency refers to each pixel's unique quality in an image.
Several gradient based methods have been presented in recent years to explain the prediction of deep neural networks \cite{simonyan2014deep,sundararajan2017axiomatic,shrikumar2017learning}.
The basic idea of gradient based explanation techniques is that each gradient measures how much a change in each input feature would change the model's predictions in a small neighborhood around the input.
Like many XAI approaches these attempt to derive an explanation from a model after it is trained, while our approach observes the changes in the model during the training process. 

Class Activation Maps (CAMs) first presented by \cite{zhou2016learning}, are another common XAI approach for deep neural networks.
The idea is to demonstrate the discriminative regions of an image used by a Convolutional Neural Network (CNN) to predict the class of the image. 
Grad-CAM \cite{selvaraju2017grad} is a common XAI approach for deep neural networks focused on identifying discriminative regions to predict the class of an image.
Grad-CAM, in addition to other methods that visualize feature interactions and feature importance like LIME \cite{ribeiro2016should} and SHAP \cite{lundberg2017unified} are the most popular current XAI methods.
We compare to these approaches in our human subject study.

There are XAI approaches, referred to as example-based XAI, that study explainability of ML models through the lens of their training data, as we do.
Using training data in machine learning explainability was pioneered by \cite{koh2017understanding}. 
They used a classic technique from robust statistics called Influence Functions to relate a model’s prediction to its training data.
Influence functions are but one of many example-based XAI approaches 
\cite{yeh2018representer,khanna2019interpreting,ghorbani2019data,barshan2020relatif,chen2021hydra}.
Hydra is an approach which studies models through the lens of their training data \cite{chen2021hydra}. 
They introduce a technique for approximating the contribution of training data by differentiating the test loss against training data weights.
Yeh et al. proposed a method to efficiently decompose the prediction (activation) value into a sum of representer values \cite{yeh2018representer}. They then select representer points – training points with extreme representer values – that could help the understanding of the model’s prediction.
However, these approaches attempt to identify something objectively true about a training set, in many cases attempting to approximate a leave-one-out training process, in which each example of the training data is left out and the model retrained from scratch.
Instead, we only consider how the final weights of a particular model emerged.
We ignore other alternative training possibilities since we want to reflect the information hidden inside the actual final weights that arose from a specific training run.
The majority of prior example-based XAI approaches require a trained model, access to the training data, and an additional evaluation process or series of metrics to determine the most important features in the training data.
In comparison, we only require access to the model during training and the training data.
To the best of our knowledge, ours is the first example-based XAI approach that derives explanations via an inspection of the specific knowledge learned during training.

The end goal of an XAI system is to provide useful intuition on an automated decision to end users. The standard way to evaluate an explanations' usefulness is to run a human subject study \cite{doshi2017towards}. Many user studies for XAI evaluation have been presented by the research community in recent years \cite{smith2020no,shen2020useful,buccinca2020proxy,chu2020visual,feng2019can,alqaraawi2020evaluating,kaur2020interpreting}. 
However, how to best conduct XAI user evaluations is still an open problem \cite{anjomshoae2019explainable,van2021evaluating}. 
Open source packages also have improved the ability to compare different XAI methods \cite{alber2019innvestigate}.
There has been also several algorithmic frameworks proposed by the researchers in recent years to evaluate XAI systems quantitatively. \cite{kim2021sanity,plumb2020regularizing,pruthi2022evaluating}. 
Despite all of this, the majority of modern XAI approaches lack formal evaluations to determine their benefit or utility to end users \cite{liao2020questioning}.
We present a novel quantitative evaluation based on evaluator accuracy and present a large-scale human subject study comparing four existing XAI approaches to Responsibility.

\section{Responsibility}

In this section, we define our Responsibility approach. 
We aim to identify the most \emph{responsible} training example for a particular decision by a machine learning (ML) model during inference.
For the purpose of simplicity, we focus on deep neural network models for this description and paper, but note that the same concepts can be applied to any ML models with parameters that individual training examples impact during the training process. 
We define the most responsible training example as the example that most altered the most activated weight. 
By most activated weight, we indicate the weight with the largest activation magnitude during inference.

Our approach relies on the introduction of an observation process during the training process.
During training, we observe the degree to which each training example alters each weight of our DNN. 
To track this change, in every batch we calculate $\Delta w_i$, or the magnitude of the change of each weight $w_i$ during that batch (${w_i}'$ is the weight after the batch) as in Equation \ref{eq2}. 
\begin{equation} \label{eq2}
\Delta w_i  = |{w_i}' - w_i| 
\end{equation}

\noindent
For each weight, we include an array ($R$) equal in size to the set of training data ($n$). We add this $\Delta w_i$ value to the index ($m$) of this array associated with each example of the the training data present in that batch. Trivially, when the batch size is equal to one, this update is equivalent to Equation \ref{eqR}.

\begin{equation} \label{eqR}
R_i[m]  +=  \Delta w_i
\end{equation}

\noindent
When the batch size is not equal to one, we instead iterate over every index ($m$) associated with that batch as in Equation \ref{eqR}.
Essentially, this means we assume that all members of a batch are equally responsible for any alteration in the weights. 
This is an assumption reliant on the choice of optimizer, and may not always be appropriate.

At times the change in weights may have a negative value. 
However, in this work we only consider the magnitude of the change, as we only wish to identify the most impactful examples.
However, we note that this might not be the case.
We have experimented with separately tracking positive and negative changes in two separate arrays ($R_pos$ and $R_neg$), which we believe has potential for regression problems. 

Consequently at the end of the training time, we will be able to identify the index $m$ of our training examples which maximally altered weight $i$ via Equation \ref{eq4}, which we refer to as the \emph{most responsible training example} for this weight $resp_i$. 

\begin{equation} \label{eq4}
resp_i = max_m(R_i)
\end{equation}

\noindent
At inference time, we can then identify the maximally activated weight, either for a particular layer or across the entire network \cite{erhan2010understanding}. 
The choice of whether to focus on a particular neuron, layer, or the entire network is an important implementation detail. 
For now, we leave this up to the particular problem domain (i.e. focusing on the neurons associated with a particular class for a classification problem). 
Once we have determined what neuron(s) to focus on, we can then identify the maximally activated weight by inspecting the output, which is dependent on the activation function of that weight.  
The output of a particular weight can be computed using Equation \ref{eq1}. 

\begin{equation} \label{eq1}
y_i = g(x_i, w_i)
\end{equation}

\noindent
where $y_i$ is the output of the weight and $g$ is the activation function. 
Since the final prediction is a complicated combination of many different weight activations, we can also consider several ``important'' weights with relatively high activations for a given prediction.
For example we can look at the distribution over activations and select the outlier values. 
Considering that we only take into account the most important weight $i$, we can then find $argmax(y_i)$ from Equation \ref{eq1} across all weights $i$ for a particular layer or for the entire network.
We can then identify the $resp_i$ value for this weight $i$, which we define as \emph{the most responsible training example}, since it is the training example that most positively impacted this weight during training. 

Unlike other approaches, which look for characteristics of the training data outside of their impact on a model \cite{lundberg2017unified}, our approach focuses on the particular training process of a particular model. 
This means that one should identify different sets of most responsible examples every time they train a model with the same training data. 
We demonstrate some relevant observations and experiments in the technical appendix.
We view this as a strength of the approach, since this means our method reflects the history of how one attained a particular model.

\section{Quantitative Evaluation}

Our goals for Responsibility are that it be general to input data representations and problem domains.
As such, as an initial test, we employ a quantitative evaluation utilizing a secondary model. 
How to best quantitatively evaluate an XAI approach is currently unclear \cite{kim2021sanity,plumb2020regularizing,pruthi2022evaluating}. 
Given that one of the purposes of XAI approaches is to improve human understanding, in this section we present a novel quantitative evaluation that attempts to approximate this process.

First, we assume that we have an ML model for a classification problem, which we refer to as the actor.
We limit ourselves to classification problems in this section as it makes evaluation simpler and due to the popularity of classification approaches in XAI literature. 
However, prior work has demonstrated Responsibility's applicability to non-classification problems \cite{khadivpour2020explainability}.
We make use of both image classification and text classification problems as an initial demonstration of the generality of the approach.
For our evaluation framework we assume that we have a secondary ML model that attempts to predict the binary performance (right or wrong) of a primary actor model, we refer to this secondary model as the evaluator. 
This evaluator attempts to approximate the process of a human attempting to improve their understanding of the actor model.
We can test the impact of including or excluding a variety of output XAI explanations as part of the evaluator's training data. 
If these XAI explanations improve the evaluator's accuracy, we take this to mean they have improved the evaluator's ``understanding'' of the actor, and thus an indicator that they may improve human understanding.
We share the term actor with actor-critic algorithms, but this evaluation should not be considered an actor-critic algorithm.

For the quantitative evaluation, we start by splitting our datasets into training, validation, and test sets.
We then train the actor using the training set with hyperparameters appropriate to the problem domain. 
We do not attempt to achieve the best possible results for the actor. 
In fact, it is best if the actor has relatively low accuracy as long as it is consistent. 
In this way we can best evaluate the impact of different XAI explanations on the evaluator's accuracy. 
Thus after training, there are inevitably samples in our validation set that the actor was not able to classify correctly. 
We refer to these as incorrect validation examples. 

For the evaluator, we would ideally like a balanced dataset, with an equal number of cases in which the actor is correct and incorrect. 
Therefore, we collect all of the incorrect validation examples and sample an equal number of correct validation examples. 
We do not attempt to sample based on the classes in the actor's classification problem, since we focus on the simpler, binary question for the evaluator of whether the actor is right or wrong. 
We call this new dataset the baseline evaluator training set. 

The baseline evaluator training set allows us to determine the accuracy of the evaluator without any XAI explanations.
We can then employ a variety of XAI approaches to produce additional explanations to augment the data of this training set. 
We concatenate each baseline training set example with its associated XAI explanation. 
We call these datasets the <blank> evaluator training set, where <blank> is replaced with the name of the relevant XAI approach.
Notably, this only works in the case where the XAI approach can output explanations appropriate as the input to a DNN. 
But given that we focus on example-based approaches, this is trivial. 
If we find an XAI approach that improves evaluator accuracy, it may even make it possible to improve the accuracy of the actor, representing a direct benefit.

We make use of three XAI approaches for this evaluation. 
The first is our Responsibility approach, implemented as we describe above for each problem domain. 
Second, for the image classification problem domain, we include Influence Functions, based on the original implementation \cite{koh2017understanding}.
We do include influence functions for our second, text classification domain given that Influence functions were not originally designed for classification. 
Finally, we include a nearest neighbor (NN) baseline.
One potential concern for Responsibility is whether the most responsible training examples captures the reasoning of the model, or whether they simply represent the most similar training example. 
Thus we include this NN baseline.
For this baseline, we determine the training example that is the closest to each example in the baseline evaluator training set of the predicted class. 
As a measure of similarity we employ Mean Squared Error (MSE) which is one of the most recognized distance measures for images \cite{palubinskas2017image}, while also working for other representations. 
For the NN baseline we consider each example of the baseline evaluator training set, given the actor model's predicted class for an example. 
We find the training example of the actor's predicted class that is most similar to the given baseline evaluator example according to MSE.
With our baselines we are specifically looking at what we can do without additional secondary processes after the network has been trained, thus we avoid approaches like Deep K-nearest neighbors \cite{papernot2018deep}.

We train three evaluator models for each problem domain, each based on one of the three datasets and then evaluate the performance of the evaluator (its prediction of whether the actor is correct or not) on the test set. 
We anticipate that if we can train a highly accurate evaluator model, we will be able to predict the correctness of the actor model on the test set.
This evaluation allows us to determine comparatively how helpful each approach is to the evaluator learning the behavior of the actor.
This is an attempt to approximate comparatively how much each of the three XAI approaches would help a human end user understand and predict the behavior of a model.
However, we do not anticipate this will be sufficient to say which approach humans would benefit from the most. 
Thus we also include the results of a human subject study below. 

We demonstrate this quantitative evaluation in the image classification problem domain with the well-known CIFAR-10 dataset \cite{krizhevsky2009learning}.
In order to include a natural language processing (NLP) task, we also run this evaluation with the PL04 dataset \cite{pang2004sentimental}.

\subsection{CIFAR-10 Problem Domain}

We include the CIFAR-10 problem domain \cite{krizhevsky2009learning}, as it is a well-understood image classification problem and dataset.
We split the dataset into 30K training, 20K validation, and 10K test sets.
For our actor we employ a four layer CNN with max pooling and dropout, followed by two fully connected layers.
We employ relu throughout except for a final softmax layer, categorical-crossentropy loss, the SGD optimizer with a learning rate of 0.001 and a batch size of one. 
This allows us to most easily determine the most responsible training example.
We train a second, simple one layer CNN as an evaluator. We use the same setup as the actor, except that we employ the adam optimizer with a learning rate scheduler decreasing from 0.0001 to 0.00001, and a batch size of 32.

\subsection{CIFAR-10 Results}

\begin{table*}[t]
\centering
\begin{tabular}{lcccc}
\hline
 & Responsibility & Influence Functions & Nearest Neighbor & Baseline \\
 \hline
Accuracy & \textbf{0.641} & 0.612 &  0.582 & 0.586  \\ 
Precision & \textbf{0.656} & 0.620 & 0.574 & 0.585 \\
Recall & 0.583 & 0.584 & \textbf{0.638} & 0.590 \\
F1 Score & \textbf{0.619} & 0.602 & 0.604 & 0.588 \\
\hline
\end{tabular}
\caption{The CIFAR-10 domain performance metrics for the different evaluators.}
\label{tab:CIFAR10}
\end{table*}

In Table \ref{tab:CIFAR10} we include the performance metrics of our evaluators over the test set.
Our Responsibility evaluator agent outperforms the other approaches by roughly 3\% in this domain. 
This demonstrates a small but clear improvement when using the responsible training examples over nearest neighbor, influence functions or no extra information.
Of particular interest is that the baseline without any extra information outperformed NN in terms of accuracy and precision. 
This suggests that the NN approach is actively harmful to the evaluator, which follows from the fact that it does not reflect any information from the actor model outside of the predicted class. 
We note that the actor's accuracy on the test set is only ~60\%.
While this is low for CIFAR-10, we expect this is due to the reduction in training data. 
Further, as stated above, it is not our goal to produce a highly accurate model.
As we demonstrate below, Responsibility does not typically hamper model accuracy.

\subsection{PL04 Problem Domain}

XAI approaches are much less commonly applied to Natural Language Processing  (NLP) problems, and even more rarely applied to both image classification and NLP problems.
Thus, we also implemented this quantiative evaluation setup for the PL04 dataset \cite{pang2004sentimental}.
The dataset is comprised of 2000 movie reviews equally divided into positive and negative semantic classes. 
We split our data into 1700 training, 150 validation, and 150 test sets, drawing equally from both classes for all splits.

We deploy a conventional CNN-LSTM model for sentiment prediction for both the actor and the critic models.
This represents the traditional model for this dataset \cite{camacho2017role}.
The only variation is that we employ a batch size of one to best identify the most responsible training example.
We evaluate our method according to a 10-fold cross-validation scheme on the dataset, and found that we performed equivalently to the original work.
We do not include Influence Functions for this problem domain as the original implementation was not designed to work with text data.

\subsection{PL04 Results}

Table~\ref{tab:PL04} shows our results for the PL04 dataset.
Across the performance metrics our responsible critic generally outperforms the two other approaches by roughly 10\%. 
Once again we note the same effect from the CIFAR-10 domain where the nearest neighbor approach led to worse accuracy. 
However, the NN does outperform the other two approaches for precision. 
We anticipate this is due to the fact that the domain only had two classes, and so is much simpler than the CIFAR-10 domain. 
Since the PL04 dataset only has two domains, we can treat the accuracy of the critic as roughly equivalent to it's accuracy on the actor's classification task. 
This means that the critic actually marginally improves over the performance of the actor (0.733 vs. 0.747). 
While this is a fairly minor improvement, it demonstrates a further utility of Responsibility.

\begin{table}[tb]
\centering
\begin{tabular}{lccc}
\hline
 & Responsibility & Nearest Neighbor & Baseline \\
 \hline
 Accuracy & \textbf{0.747} &  0.626 & 0.686  \\ 
 Precision & 0.762 & \textbf{0.804} & 0.722 \\
 Recall & \textbf{0.974} & 0.661 & 0.658 \\
 F1 Score & \textbf{0.855} & 0.725 & 0.689 \\
 \hline
\end{tabular}
\caption{The PL04 domain performance metrics.}
\label{tab:PL04}
\end{table}

Comparing the CIFAR-10 and PL04 results, PL04 seems better suited to our approach. 
However, we anticipate this is largely due to the size of the dataset, models, and complexity of the problem (10 vs. 2 classes). 
Thus, we do not think that Responsibility is generally more suited to NLP problems.

\section{XAI Comparison Human Subject Study}

\begin{figure*}
    \centering
    \includegraphics[width=6in]{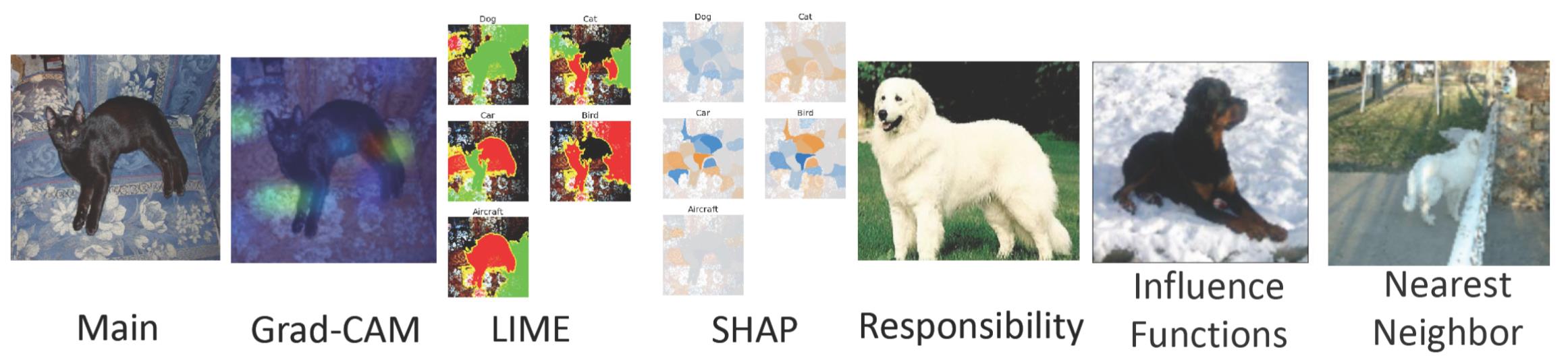}
    \caption{Example of the approaches in this study. The three existing XAI approaches (left), our Responsibility approach, Influence functions approach, and the Nearest Neighbor baseline (right).}
    \label{fig:comparisonFigure}
\end{figure*}

We ran a human subject study on Amazon Mechanical Turk (AMT) to compare between our Responsibility XAI approach and existing XAI approaches.
We chose AMT as our focus is on non-expert end users, and due to the number of subjects required given the size of this study. 
To compare the utility of these different approaches to non-expert users, we focused on the task of users guessing the prediction of an image classification model.
We chose an image classification task as the majority of XAI work has been for the image classification problem domain, which would thus make a comparative study simpler. 
We chose to have users attempt to predict the classification of the model as a simple approximation for how well users understood the model's behavior from the XAI explanation.

Since our problem area was image classification, we made use of four popular XAI approaches for classification tasks.
They were Local Interpretable model-agnostic explanations (LIME) \cite{ribeiro2016should}, Shapley additive explanations (SHAP)  \cite{lundberg2017unified}, Grad-CAM \cite{selvaraju2017grad}, and Influence Functions \cite{koh2017understanding}.
We chose this set as all four represented general approaches with distinct visualizations. 
We focused on distinct visualizations so that non-expert humans could differentiate between the explanations.
Thus, we did not include any variations of these approaches, which might have required XAI expertise to differentiate between.

In addition to these four existing XAI approaches we also include the Nearest Neighbor (NN) baseline from the quantitative evaluation.
For a given input image, this baseline selected the training example from the model's predicted class that minimized the MSE loss as before.
Thus, this was a strong baseline, essentially informing the user of the correct class, but without any real basis in what the model learned.
However, Responsibility and Influence Functions should also inform the user of the correct class, while ideally also reflecting the trained model.

We include an example visualization of all the XAI visualizations for a given test image in Figure \ref{fig:comparisonFigure}, with the input test image on the left. 
The correct class is cat, but the predicted class is dog. 
The Responsibility image appears to clearly convey the predicted class, but may seem somewhat unusual or lacking similarity to the original test ``main'' image.
However, the dark background of the Responsibility image does share some similarity with the dark cat. 
Considering the LIME visualization, we can confirm that it is the case that model is paying attention to the shared darkness between the two images. 
We found this type of shared characteristic common for the most responsible examples compared to particular input images. 

\subsection{Dataset and model}

To run our user study we needed a trained ML model to compare the quality of explanations generated by different XAI methods.
We made use of the VGG-16 architecture introduced in \cite{simonyan2014very}, as a standard and high quality image classification model. 
We could not use CIFAR-10 as our dataset.
Many popular image classification datasets have too low of a resolution for non-expert human judgement to be meaningful, such as CIFAR-10.
As such, we constructed a novel dataset by combining five image datasets with relatively high resolutions. 
These were the Caltech-UCSD Birds-200-2011 \cite{wah2011caltech}, CAT \cite{zhang2008cat}, Stanford Dogs \cite{khosla2011novel}, FGVC Aircraft \cite{maji2013fine}, and Stanford Cars \cite{krause20133d} datasets.
We trained VGG-16 as a classifier on these five image classes. 
We used the same training setup as in the CIFAR-10 quantitative evaluation in order to identify the responsible training examples.
For the human subject study, we only made use of test images and their associated explanations. 

\subsection{Methodology}

We employed a within-subjects, comparative evaluation due to the benefits of ranking evaluations in human subject studies in comparison to rating evaluations \cite{martinez2014don}. 
Thus, every human subject was randomly assigned two of our XAI approaches when they accessed the study website. 
The subjects were non-experts, and so were given a short tutorial on the definition of image classification and explainable artificial intelligence. 
They were then introduced to their two assigned XAI approaches with two example test images (one correct, one incorrect) and the associated explanations from both approaches. 
These introductory examples were fixed, such that all participants assigned to the same XAI approach saw the same tutorial input images and visualization.
The assignment of the pair of XAI approaches was done at random.

After the introduction, the main experiment began.
Each participant first was given five test images alongside the explanation visualizations of their two assigned XAI approaches. 
For each of these first five test images, users were informed of the correct class and the model's prediction.
After these initial five ``training'' examples where the participants were given the models' predictions, they were presented with five new test images along with the associated explanation visualizations. 
However, for these second five examples participants were not given the model's predicted class. 
Instead, participants were asked to pick which of the five classes they thought the model predicted.
The model's prediction could be correct or incorrect, and the participants only had their experience and the pair of explanations to guide their decision.
Thus the first five examples acted as a training set and the second five acted as a test set. 
Finally, participants were asked to rank the two XAI approaches in terms of preference and understandability.

Each participant was guaranteed to see a unique ten test images, determined by their study ID. 
We made this choice to minimize the risk of measuring differences between particular images instead of between the XAI approaches. 
This was also done to ensure AMT workers could not collude on the correct model predictions. 
Each participant had an equal number of examples in which the model was correct and incorrect spread randomly across their examples. 

Participants were given a maximum of fifteen minutes for the study, though all participants finished in less than ten minutes.
Each participant was paid 3.00 USD for taking part and an addition 0.20 USD for each correct answer for the second five examples.
This was meant to incentive AMT workers to try to guess correctly and to combat the `carelessness' problem, in which AMT workers attempt to complete tasks as quickly as possible to maximize their income~\cite{aruguete2019serious}.
This study was approved by a Research Ethics Board at the University of Alberta, ethics ID number (Pro00110990).

\subsection{Results}

\begin{table*}[t]
\begin{tabular}{lcccccc}
\hline
 & Responsibility & Grad-CAM & LIME & SHAP & Influence Functions & Nearest Neighbor \\
 \hline
More Understandable & \textbf{59} & \textbf{59} & 50 & 36 & 46 & 50 \\
More Preferred & 53 & \textbf{59} & 45 & 47 & 42 & 51 \\
 Pair Accuracy & 62.6$\pm$3.7 & 54.8$\pm$8.9 & 56.2 $\pm$11.5 & 55.4$\pm$9.0 & 50.2$\pm$7.9 & \textbf{64.8$\pm$7.7} \\
 Preference Accuracy & \textbf{83.64} & 47.47 & 50.0 & 53.33 & 42.72 & 81.50\\
\hline
\end{tabular}
\caption{Overview of the AMT study results.}
\label{tab:StudyResults}
\end{table*}

We collected results from 300 AMT workers.
This meant that we collected a total of 3000 predictions, or 100 predictions for each XAI approach. 
We first ran a multi analysis of variance (MANOVA) to determine whether the ordering of the XAI approaches or any of the other demographic information impacted any of the results. 
The MANOVA found no significant impact, and so we can safely treat this data as coming from only 15 conditions: the 15 distinct pairs of XAI approaches. 
Thus, we had 20 participants in each condition, more than sufficient for statistical analysis. 

We include the overall quantitative results in Table \ref{tab:StudyResults}, marking the largest value in each column in bold. ``More Understandable'' is the total number of times that, that approach was chosen as more understandable by a participant out of the 75 participants who saw that approach. 
This number sums to 300. 
We found that Responsibility and Grad-CAM were the two approaches most commonly chosen to be more understandable.
Notably this is summing over all possible pairs of XAI approaches.
With a paired Wilcoxon Mann Whitney-U test, chosen due to a lack of a normal distribution, we found that LIME was considered significantly more understandable than SHAP, and Responsibility was considered more understandable than LIME $p<0.005$.
However, there was no significant difference between SHAP and Grad-CAM on understandability, indicating that the two approaches were too similar for participants to differentiate. 

``More Preferred'' indicates the number of times, out of the 75 participants who saw that approach, that it was overall preferred. 
Grad-CAM is the clear winner here, though Responsibility comes second. 
There are no significant results in terms of overall preference.

We include two different accuracy terms, both concerned with the rate at which participants identified the class the model predicted for the second five examples.
The first, ``Pair Accuracy'' is the average accuracy taken across all five pairs that include each approach. 
We employ this due to the complexity of determining whether the participant's performance comes from one of the XAI approaches presented or from a combination of both. 
NN outperforms all other approaches according to this measure. 
This may seem somewhat trivial since NN explicitly informed the user of the model's chosen class. 
However, both Responsibility and Influence Functions also explicitly informed the user of the model's chosen class (as in Figure \ref{fig:comparisonFigure}).
The only significant difference is that both Responsibility and NN are significantly different than Influence Functions.
This shows a clear distinction between Influence Functions and Responsibility. 

``Preference Accuracy'' is the measure of the average accuracy of the participants at choosing the class the model predicted when they indicated that, that XAI approach in question was their preference (``More Preferred'').
We included this to get around the same pair issue as above, taking participants at their word that their selected approach was the one they relied on.
This gives a different result, with Responsibility taking the lead, NN performing very similarly, and all other approaches having accuracy at around the level of chance. 
We anticipate this is due to participants believing that they understood an explanation per the ``Most Understandable'' results, but not being able to identify the predicted class based on it.
If class was the major factor, Influence functions should have performed well here, but participants seemed to have a harder time understanding how the training data example related to the test example. 

Overall, we take these results to indicate that Responsibility differs from existing XAI approaches and may be more beneficial to human understanding.
These results seem to validate our proposed quantitative evaluation, demonstrating that improving evaluator accuracy does appear to approximate improving human accuracy.

\section{Discussion}

In this article we have introduced Responsibility.
We claim this approach is general, and can be used across different problem domains and models, and that it is beneficial to end users. 
We demonstrate this through a series of experiments, including a human subject study and a quantitative study comparing Responsibility to other XAI approaches across two different domains.

Our quantitative results demonstrated that Responsibility benefited a secondary evaluator DNN model in predicting the behavior of a primary actor DNN model across two domains. 
However, the human subject study results complicated this somewhat. 
While the Nearest Neighbor visualizations actively harmed the performance of our evaluators, humans seemed to benefit from them. 
Despite this, the Nearest Neighbor visualizations were not ranked as more understandable than any other method. 
Thus, our interpretation of these results is that the Nearest Neighbor visualizations did inform users of the correct class, but did not help them understand the model's learned behavior. 

In terms of potential harmful implications, we hope our Responsibility approach can be used to mitigate the harm from lack of understanding in black box ML systems. 
However, there are ways that this approach could also be weaponized against end users. 
While we have evidence that this approach helps AI evaluators and human users accurately reflect the behavior of a target model, at least in comparison to existing AI approaches for the latter, this does not mean it alone is sufficient for users to understand a target model. 
Applied naively, Responsibility could even lead users to develop an overly simple understanding of a model's behavior, putting them at risk in situations where understanding the details of a model's decision making process matters.
We strongly discourage such applications, particularly without future study into Responsibility and its effects.

\section{Conclusions}

In this paper, we present a novel Explainable AI (XAI) approach we call Responsibility. 
We give a technical definition and description of the approach, and evaluate it quantitatively and qualitatively. 
We found that the approach was able to consistently improve the performance of a secondary model attempting to learn to predict the behavior of a primary model. 
We also found that non-expert humans considered the approach to be more understandable than existing XAI approaches, and that it improved user accuracy relative to four existing XAI approaches. 
Overall, we view Responsibility as a general XAI approach with the capacity to benefit both humans and machines.

\bibliographystyle{named}
\bibliography{main}

\clearpage

\appendix

\section{OBSERVATIONS AND EXAMPLES}

In this section, we present a number of observations and experiments meant to improve the reader's understanding of Responsibility.
We employed MNIST \cite{lecun1998gradient}, a dataset of hand-drawn human digits, for the majority of this section due to it's popularity and ubiquity.
We also give some examples that rely on the Fashion MNIST dataset \cite{xiao2017fashion}.
For the majority of examples we choose the LeNet-5 network \cite{lecun1998gradient} with 61,706 trainable parameters as our model, as we wanted to employ a simple but reasonably strong model. 
We trained LeNet-5 and tracked the most responsible training values as discussed in Section 4 across all layers, though we focus on the first and last layers in these examples.

\subsection{Responsibility Values}

\begin{figure}[t]
    \centering
    \includegraphics[width=0.95\columnwidth]{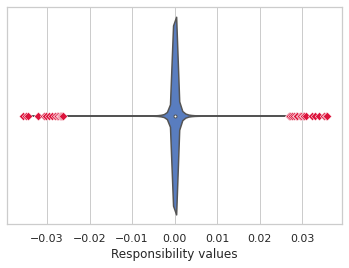}
    \caption{The violin plot demonstrates the distribution of the responsibility values using density curves. The width of the colored area represents the approximate proportion of the data located in that region.}
    \label{fig:violin}
\end{figure}

\begin{figure}[t]
    \centering
    \includegraphics[width=0.95\columnwidth]{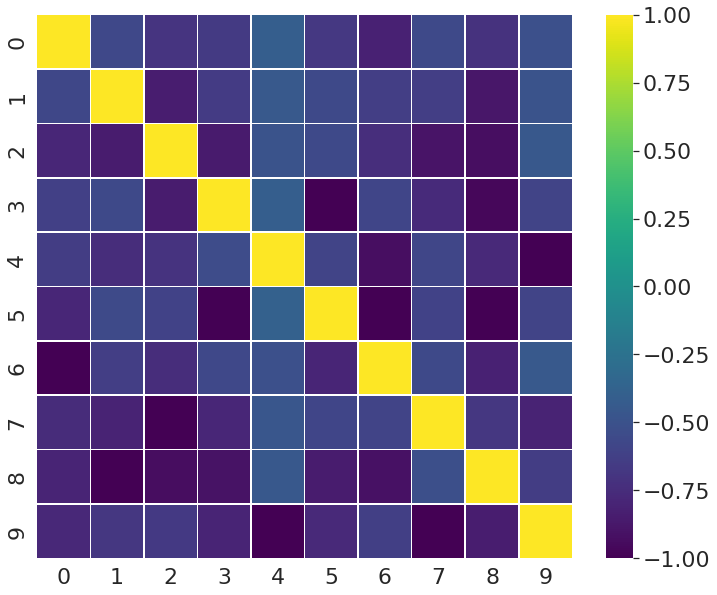}
    \caption{Inter-class responsibility shown as a heatmap. Rows represent classes of training data and columns represent the neuron associated with the particular class.}
    \label{fig:inter_class}
\end{figure}

For our first example, we compute the mean and standard deviation of the changes of all training data points of the neurons of the final layer. 
These values show how much each training data point is responsible for the changes of each neuron during the training phase. 
The mean and standard deviation are $(6.84\mathrm{e}{-10}$ and  $3.36\mathrm{e}{-4})$, respectively.
The mean value is very close to zero, and the standard deviation is substantially higher than the mean. 
You can find the distribution visualized in Figure \ref{fig:violin}. 
High width of the curve around zero depicts high frequency of data points in that area. 
The 20 maximum and minimum values also are demonstrated in red.
This follows our intuition that the contribution of a single training data point in a large dataset is likely small. 
The standard deviation and the red dots demonstrates that some data points have extreme contribution values. 
This matches our intuition that we can differentiate a small subset of individual training data points that have a large impact on the final model, which supports the basic concept of Responsibility.

\subsection{Inter-class Responsibility}

Next, we inspect the overall impact of particular classes in the training data in terms of the particular neurons used for classification.
We expect each training data example alters all neurons of the last layer during training phase, but each training data example has a very different impact in terms of the degree of change.
We sum up these changes and separate them with respect to the labels of the training examples (0-9, since MNIST is a dataset of hand-drawn digits). 
We plot this inter-class Responsibility in Figure \ref{fig:inter_class}. 
The rows represent training data classes and the columns represent the neuron associated with a particular class (0-9). 
We would anticipate that training data points of the same class would have the largest impact on the neuron associated with predicting that class, which Figure \ref{fig:inter_class} clearly supports.

The maximum Responsibility values for each row and column appear on the diagonal, indicating that training data of a particular class always makes the largest changes in neurons representing the same class. 
We observe symmetrically low values between the digit pairs (3,5), (2, 7), (4,9), and (7,9). 
As the two classes in each pair have similar appearances, this likely is due to competition between them.
In other words, training data points representing the digit ``5'' decrease the parameter values of the neuron associated with predicting ``3'', making it less likely for the model to predict ``3'' when given a 5. 
Therefore, by visualizing Responsibility, Figure \ref{fig:inter_class} represents a visualization of the model's training process. 
Specifically, we can identify darker cells as cases where the model was more often incorrect during the entire training process and brighter cells as cases where it was more often correct.

\subsection{Data debugging}

\begin{figure*}[t]
    \centering
    \includegraphics[width=0.9\textwidth]{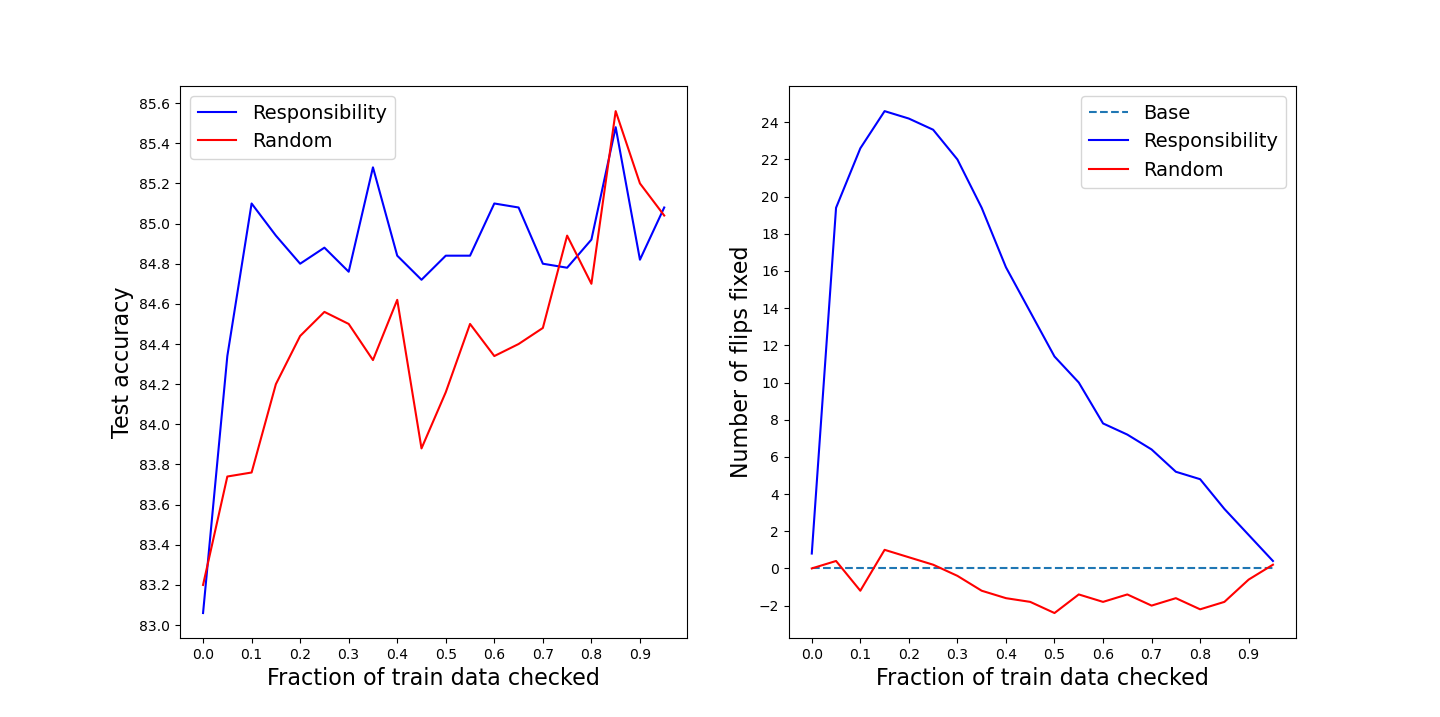}
    \caption{Graphs comparing the impact of using Responsibility to determine the order to inspect and correct mislabeled data in comparison to a random order.}
    \label{fig:data_debugging}
\end{figure*}

In this example we investigate the utility of Responsibility applied to data debugging, or to identifying mislabeled training data. 
We expect mislabeled examples to be among the most responsible data points, due to the fact that they should stand out as distinct from other data points of the same class. 
By the most responsible data points, we mean training samples which have the largest impact on the neurons of the first layer during training. 
We calculate this impact by considering the absolute changes of the neurons in the first layer. 
We focus on the first layer over the final layer here as we anticipate the majority of the discriminatory features to be found in the first layer. 

We employ 1000 randomly sampled images from the Fashion MNIST dataset \cite{xiao2017fashion} to classify T-Shirts and Tops against Shirts.
We randomly flip the label for 10\% of our training points to produce a training dataset with mislabeled points. 
We require a simpler model for this example as it involves retraining the neural network a considerable number of times, and we also wanted to ensure that the model would not be robust to mislabeled data.
As such, we train a convolutional neural network with one convolutional (8 kernels of size 3 by 3) and two fully connected layers (sizes of 128 and 2, where 2 is our two classes) on this noisy dataset.
We then order the 1000 training datapoints according to their Responsibility.
In comparison, we randomly order the 1000 training datapoints. 
We then iterate through both sequences, representing a manual check of each of the 1000 datapoints. 
When we reach a particular datapoint, we correct it's label if it was incorrect and then retrain the model.
Our goal would be to minimize the amount of manual checks required. 
Figure \ref{fig:data_debugging} visualizes this process. 
The left graph gives the test accuracy on the Fashion MNIST test set for our model during the data debugging process for the Responsibility and Random sequences.
By moving from left to right, it is shown that the test accuracy is improving while flipping a higher fraction of incorrect labels in both sequences. 
However, as is shown, using Responsibility we can have a higher test accuracy in early stages of the data debugging process.
The right graph gives the number of mislabeled data points that have been fixed at each sequence compared with the ideally random baseline. 
For example, by checking 0.2 of the data points we find 24 more mislabeled data points in the Responsibility sequence. 
It is clear that using Responsibility we can identify mislabeled data points sooner than with a random inspection.
As a result we can have a higher test accuracy in early stages of data debugging using responsibility compared with the random debugging, saving hypothetical human time and effort.   

\subsection{Understanding Misclassified Examples}

\begin{figure*}[t]
    \centering
    \includegraphics[width=0.9\textwidth]{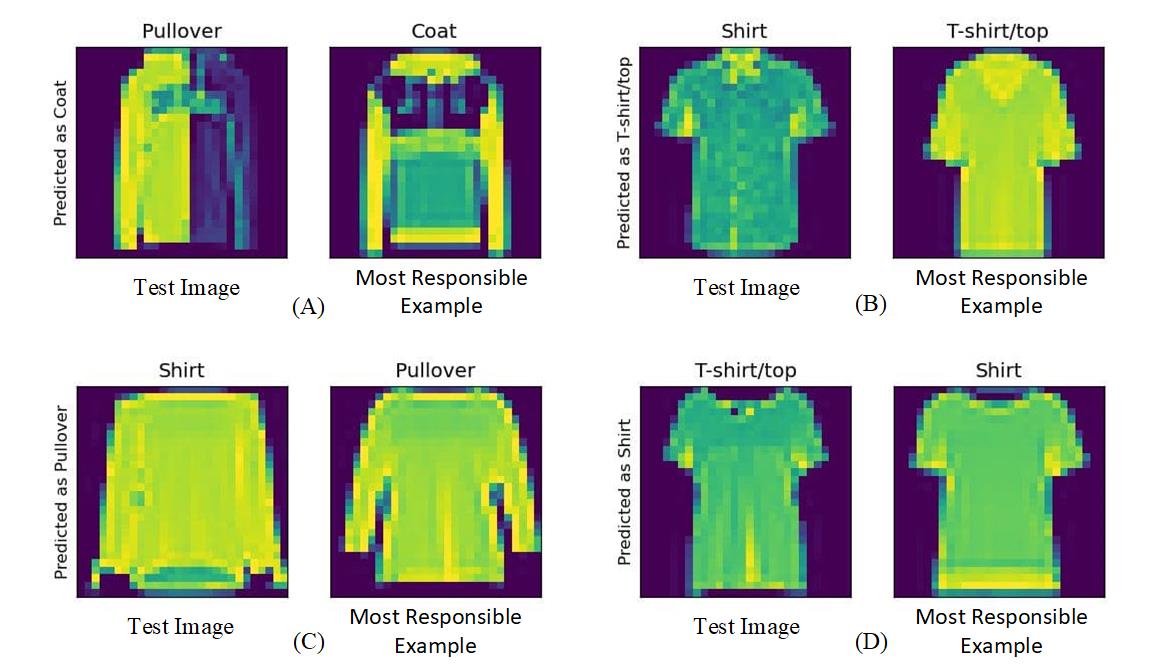}
    \caption{A visualization of four randomly selected test samples and their associated most responsible training example.}
    \label{fig:misclass}
\end{figure*}

In this example, we consider how Responsibility can help us better understand classification errors. 
This time, we make use of the entirety of Fashion MNIST, and the same model as in the prior example except with the final layer expanded to account for all ten classes. 
Fashion MNIST dataset makes it possible for us to have a better understanding of the misclassified examples, and for this particular experiment, it is more helpful compared with the MNIST dataset. 
Thus we decided to use this dataset in this experiment.
After training, we used the model to evaluate 600 of the test images, equally sampled across all ten classes. 
Even with such a simple model, the total number of misclassified examples is only 70. 
We visualize four of these 70 examples (left) and their most responsible training example (right) in Figure \ref{fig:misclass}.
Notice that the most responsible examples have similar appearances to the test example even though they are belong to another class. 
Pairs (C) and (D) are particularly close, with nearly the same color and silhouette. 

Example (A) in Figure \ref{fig:misclass} is one of seven pullovers misclassified as a coat. 
Of these seven, six share the same most responsible training image. 
A similar pattern can be identified for the other three examples. 
Example (B) is one of the seven shirts misclassified as a T-shirt, with five of these sharing the same most responsible image .
Example (C) represents an example of one of the three shirt that are misclassified as a pullover, with the given most responsible training image being the same for all three. 
Finally, Example (D) represents one of the seven T-shirts misclassified as a shirt. It's most responsible training example is shared with five of these seven. 
This demonstrates how Responsibility can give us an insight in terms of where our model fails and the kinds of classification problems where it will struggle.
This could also help in terms of removing training samples that might confuse or negatively impact a model, and may also indicate that these examples would confuse humans as well.

\subsection{Responsible Examples Across Different Seeds}

\begin{figure*}[t]
    \centering
    \includegraphics[width=0.65\textwidth]{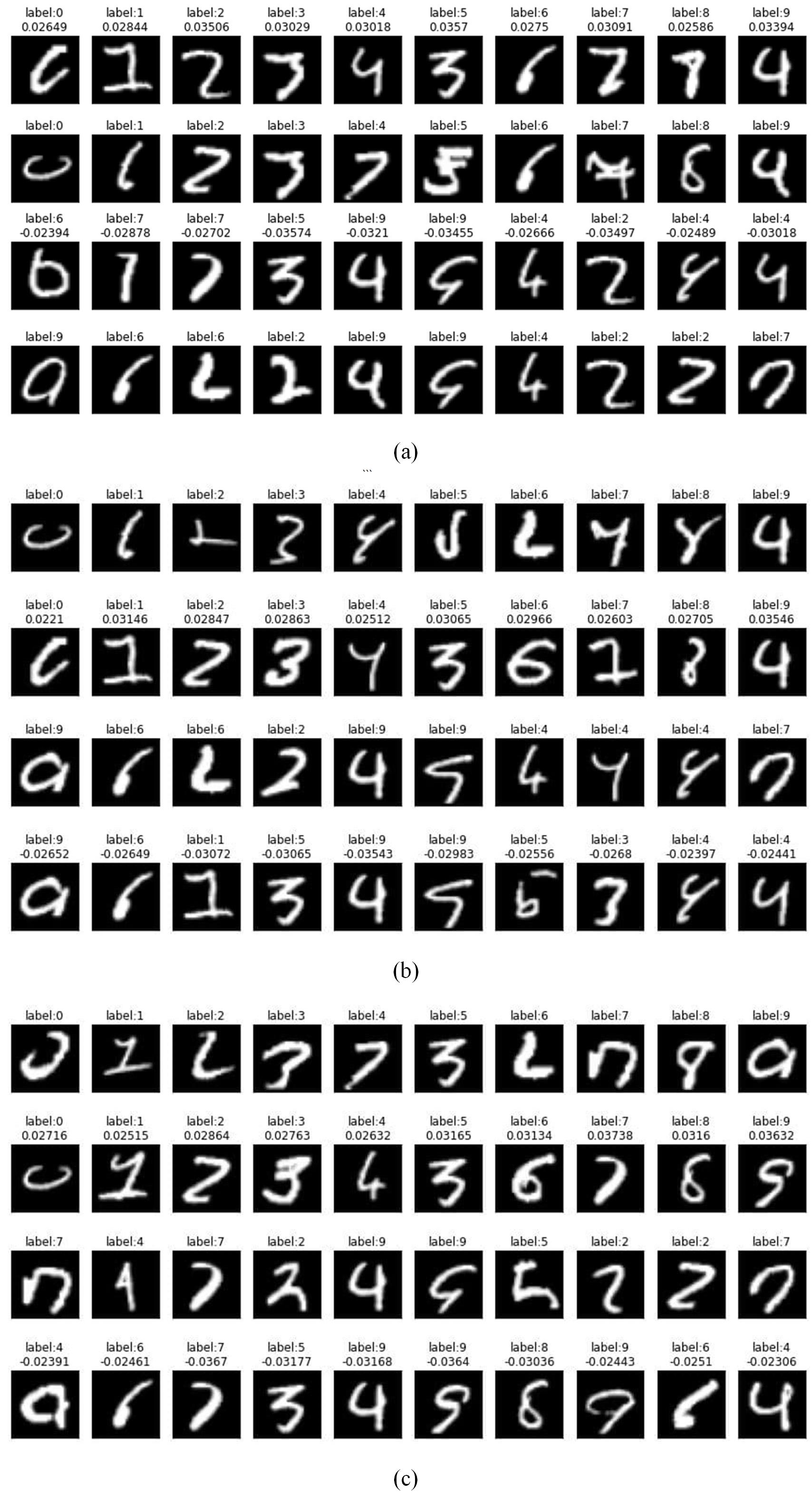}
    \caption{The most responsible examples of the training data across different seeds. (a): Random seed 0, (b): random seed 1, (c): Random seed 2. 
    \newline First rows: Data points which most frequently identified as the most responsible (Positive change in the neurons).
    \newline Second rows: Data points with the largest positive change in the neurons.
    \newline Third rows: Data points which most frequently identified as the most responsible (Negative change in the neurons).
    \newline Second rows: Data points with the largest negative change in the neurons.}
    \label{fig:seeds}
\end{figure*}

One of our primary assumptions with Responsibility is that different training processes will lead to distinct final models, and that XAI approaches should capture these differences. 
In this final example we identify Responsibility data points across three different seeds for the MNIST dataset and the LeNet-5 model.
We train the model three times with three different random seeds. 
Figure \ref{fig:seeds} shows these Responsibility data points across three different random seeds. 
The first rows give the most frequent training data points that make positive changes in the output neurons.
The second rows are the training data points which make the largest positive change in the output neurons. 
The third rows are the most frequent training data points that make negative change in the output neurons.
The fourth rows are the training data points which make the largest negative change in the output neurons. 
Thus, each figure has a total of forty images, corresponding to different ways of measuring Responsibility for the ten final classification neurons of the model across three different random seeds. 

The first and second rows of Figures \ref{fig:seeds} show that the data points with positive contributions tend to have the same labels as the target labels. 
However, they tend to have unusual appearances. 
This matches prior example-based results in terms of helping to identify outliers in the training data.
The third and fourth row demonstrate that the data points with the extreme negative contributions do not have the same labels as the target labels but they closely resemble the target labels.
While there are shared training examples across all three figures, each also has a unique set. 
This confirms that Responsibility can capture differences in how each model was trained, even subtle ones like the initial Random seed.

\end{document}